\documentclass[conference]{IEEEtran}

\usepackage{cite}
\usepackage{amsmath,amssymb,amsfonts}
\usepackage{algorithmic}
\usepackage{graphicx}
\usepackage{capt-of}
\usepackage{textcomp}
\usepackage{xcolor}
\usepackage{booktabs}
\usepackage{multirow}
\usepackage{siunitx}
\usepackage{stfloats}
\usepackage{placeins}

\usepackage[hidelinks]{hyperref}
\usepackage{orcidlink}

\begin{document}

\raggedbottom

\title{
Prompt Re-Specification and Target-Mask Availability in Mask-Conditioned
Visual Reinforcement Learning for Moving-Platform Landing
}

\author{
\IEEEauthorblockN{
Ho Tim Tsui\,\orcidlink{0009-0007-5488-0735},
Jongyun Kim\,\orcidlink{0000-0003-1698-2726}
}
}

\maketitle

\begin{abstract}
Visual reinforcement learning (RL) policies that land a drone on a moving
platform can fail when the landing pad looks different from the one used
in training. Using binary target masks as the visual input removes most
pixel-level appearance information while preserving the spatial
representation of the target, and lets a text prompt specify the target. In simulation, we test whether a frozen (fixed-weight) mask-based DrQ-v2
policy can handle a new pad when only the prompt of its fine-tuned SAM~3
detector changes. A banana-themed pad lowered landing success from 75.6\%
to 12.0\%, and SAM~3 rarely detected it under the original prompt. A
revised prompt, chosen after testing several candidates, restored success
to 68.4\% with no weight updates. In a separate experiment with a policy
trained on ground-truth masks, filling in masks that were missing while
the pad was visible helped landing far more than the tested mask-centroid
and area corrections. Tracking the pad over time supplied masks more often and
also kept producing them after the pad was confirmed absent from the
rendered image, but it did not establish a reliable improvement in
landing success. Whether a usable
mask reaches the policy at all thus appears to be an important limiting
factor for this approach.
\end{abstract}

\section{Introduction}
\label{sec:intro}

Autonomous landing on moving platforms is an important capability for
unmanned aerial vehicles (UAVs) in mobile recharging, logistics,
inspection, and maritime operations~\cite{goldschmid2024}. Cameras offer
lightweight sensing without requiring communication with the platform or
external positioning infrastructure~\cite{falanga2017,ladosz2024}, and
reinforcement learning (RL) can learn tracking, alignment, and descent
together in one closed-loop policy. Ladosz et al.~\cite{ladosz2024}
trained such a policy with DrQ-v2~\cite{yarats2022drqv2}. It maps camera
images and onboard velocity and attitude to velocity commands, without
access to the position or velocity of the platform.

A policy that learns from unsegmented camera images can become tied to the
appearance of the landing pad it saw during
training~\cite{polvara2020,ladosz2024}. If the pad is repainted or
replaced, the policy may need new data and retraining. Promptable
segmentation models~\cite{kirillov2023sam,carion2026sam3} offer another
route. A perception model finds the target from a text prompt and passes
it to the controller as a binary mask. Most of the appearance of the pad
then never reaches the policy, and a changed pad can be described by a new
prompt. This alone does not make landing robust to appearance, because the
perception model still has to detect the new pad. When direct detection
fails, the perception frontend falls back on temporal propagation, as in
video segmentation models with memory~\cite{ravi2025sam2}, and this
propagation can also fail or carry outdated target information into later
frames~\cite{ding2025sam2long}. In closed loop, the frames that the policy
sees also depend on its own earlier actions, and frame-level segmentation
accuracy need not predict downstream performance~\cite{philion2020}.

This paper asks whether a frozen mask-conditioned RL landing policy can
adapt to a new landing-pad appearance when only the segmentation prompt
is changed, without updating any network weight, and what limits this
adaptation. Throughout the paper, ``frozen'' means that the weights of a
network are not updated after its training, so a frozen network has the
same weights in every condition in which it is evaluated. We study the question in simulation with a DrQ-v2 actor that
observes masks from a SAM~3 detector fine-tuned for the task. When a
banana-themed pad replaced the training pad, direct detections and landing
success collapsed under the original prompt. A revised prompt brought back
most of the detections and most of the landing success. A substitution
diagnostic with a separately trained ground-truth-mask policy points to
the property involved. With raw SAM-derived masks, this policy almost
never landed. Filling in only the masks that were missing while the target
was visible restored landing, whereas correcting the centroid and area of
the masks that were produced did not. The perception frontend already
uses temporal tracking to supply masks through detection gaps.
Reinitialising the tracker at each direct detection made the target
available more often, but it also produced more nonempty masks after
renderer-confirmed target absence, and the resulting difference in
landing success was not established.

The central contribution of this work is an empirical account of
prompt-conditioned segmentation as the observation interface of a frozen
RL landing policy when the target appearance changes. It identifies
target-mask availability as a key limitation for control. Three connected
investigations support it. A matched appearance and prompt evaluation
(Section~\ref{sec:appearance}) shows that revising the prompt recovered
most of the landing performance lost under the tested appearance change.
A frozen-policy substitution diagnostic (Section~\ref{sec:diagnosis})
shows that, among the interventions tested, restoring missing masks
mattered far more for landing than correcting mask geometry. A comparison
of temporal memory management (Section~\ref{sec:temporal}) shows that
greater target continuity comes with more nonempty output after
renderer-confirmed target absence and does not automatically improve
landing. The results concern one substantially different pad appearance,
a prompt revised by the authors, and one trained actor, all in
simulation.

\section{Related Work}
\label{sec:related}

\subsection{Moving-Platform Landing}

Classical systems treat moving-platform landing as modular estimation and
control. Image-based visual servoing uses image-space target features and
optical flow~\cite{herisse2012,serra2016}, and feed-forward variants add an
estimate of the platform velocity~\cite{cho2022}. Falanga et
al.~\cite{falanga2017} predict the platform motion with an extended Kalman
filter while detections are unavailable, and Paris et
al.~\cite{paris2020} plan trajectories whose final velocity matches the
estimated platform motion. These systems bridge visual gaps with an
explicit state estimate, but they rely on an engineered marker and a
structured pipeline. Learning-based methods replace part or all of this
pipeline. They use extracted relative state~\cite{rodriguezramos2019},
sub-task decomposition~\cite{goldschmid2024}, or image-based policies
transferred with domain randomisation~\cite{polvara2020,ladosz2024}. The
end-to-end policy of Ladosz et al.~\cite{ladosz2024} assumes that the pad
design was seen during training, and its environment and learner are the
closest to ours. Other recent methods handle partial observability inside
the policy, with a long short-term memory network over landing-specific
keypoints~\cite{shin2026} or recurrent RL over noisy and missing state
measurements~\cite{aikins2024}. We keep the actor feed-forward and deal
with appearance and missing observations in the perception interface.

\subsection{Perception Representations for Learned Control}

The representation passed from perception to a learned controller decides
which errors the controller has to tolerate. A driving policy shielded
from raw perceptual input by a modular perception layer transferred from
simulation to a real vehicle~\cite{mueller2018}, and agents given
intermediate vision representations learned faster and generalised better
than agents learning from pixels alone~\cite{zhou2019}. In visual RL,
images masked with Segment Anything improved generalisation to unseen
visual conditions~\cite{wang2023samg}. Segment
Anything~\cite{kirillov2023sam} segments objects from visual prompts, and
SAM~3~\cite{carion2026sam3} extends promptable segmentation to concepts
given as text, so a text prompt can describe a new target without
retraining the controller. In our setting the appearance change concerns
the landing target itself, and the adaptation is made through the text
prompt. Bounding boxes and keypoints are more compact alternatives to
masks; we do not compare interface types. Because perception metrics need
not reflect downstream performance~\cite{philion2020}, we evaluate the
interface by closed-loop landing.

\subsection{Temporal Memory in Video Segmentation}

Video object segmentation carries information across frames.
SAM~2~\cite{ravi2025sam2} conditions each frame on a streaming memory of
earlier predictions, XMem~\cite{cheng2022xmem} organises memory into
sensory, working, and long-term stores, and
DAM4SAM~\cite{videnovic2025dam4sam} adds distractor-aware memory selection
to SAM~2.1. Memory can also spread mistakes:
SAM2Long~\cite{ding2025sam2long} shows that the greedy memory selection of
SAM~2 lets a wrong or missed mask affect later frames. These methods are
evaluated by segmentation accuracy on recorded video. We instead measure
how memory management changes the observations that a closed-loop
controller receives, and how landing outcomes differ.

\section{Task and System}
\label{sec:system}

\subsection{Landing Task}
\label{sec:task}

A quadrotor lands on a non-communicating moving platform using only
onboard visual and ego-state observations. The simulation environment is
built on \texttt{gym-pybullet-drones}~\cite{panerati2021}. It is adapted
from the moving-platform landing environment of Ladosz et
al.~\cite{ladosz2024} and shares its nominal start height, horizontal
reset range, and maximum platform speed. The vehicle is the
Crazyflie~2.X model of the simulator with a rigid, downward-facing camera,
and a proportional--integral--derivative (PID) controller tracks its
velocity commands. The landing target is a
\SI{1}{m}~$\times$~\SI{1}{m} pad bearing a fixed greyscale circular
landing-marker texture, which is not randomised during training.

Each episode starts with the UAV at a height of \SI{10}{m} and a
horizontal offset from the platform drawn uniformly from
$[-3,3]$~\si{m} along each axis. The platform moves along a straight line
at a speed drawn uniformly from $[0,1]$~\si{m\,s^{-1}} and a heading drawn
uniformly from $[-15^\circ,15^\circ]$. The physics simulation runs at
\SI{240}{Hz}, the PID controller at \SI{24}{Hz}, and the actor at
\SI{12}{Hz}. Fresh camera images are segmented at \SI{6}{Hz}, so the actor
acts on each mask for two consecutive decisions. An episode ends at the
first contact or after \SI{30}{s}. A landing succeeds when the UAV
touches down with its horizontal collision footprint, of radius
\SI{0.06}{m}, inside the pad. Both horizontal offsets from the pad centre,
measured in the pad frame, must therefore be at most \SI{0.44}{m}, and no
limit is placed on touchdown velocity. Any other contact is a crash, and an
episode still airborne at \SI{30}{s} is a timeout.
Table~\ref{tab:task_setup} summarises the parameters.

\begin{table}[t]
\centering
\small
\caption{Landing-task parameters.}
\label{tab:task_setup}
\begin{tabular}{lc}
\toprule
Parameter & Value \\
\midrule
Physics / PID rate              & \SI{240}{Hz} / \SI{24}{Hz} \\
Actor rate                      & \SI{12}{Hz} \\
Perception update rate          & \SI{6}{Hz} \\
Camera model                    & pinhole, no distortion \\
Focal length $f_x=f_y$          & 63.455~px \\
Principal point $c_x=c_y$       & 41.5~px \\
Field of view                   & approx.\ $67^\circ\times67^\circ$ \\
Mask resolution                 & $84\times84$ \\
Camera near plane               & \SI{0.0397}{m} \\
Landing-pad size                & \SI{1}{m}~$\times$~\SI{1}{m} \\
Initial height                  & \SI{10}{m} \\
Initial horizontal offset       & $\mathcal{U}[-3,3]$~\si{m} per axis \\
Platform speed                  & $\mathcal{U}[0,1]$~\si{m\,s^{-1}} \\
Platform heading                & $\mathcal{U}[-15^\circ,15^\circ]$ \\
Episode horizon                 & \SI{30}{s} (360 actor decisions) \\
\bottomrule
\end{tabular}
\end{table}

\subsection{Perception Frontend}
\label{sec:frontend}

The perception frontend turns each fresh camera image into the mask and
perception metadata that the actor observes. It uses three frozen models.
The direct detector is a SAM~3 image segmenter~\cite{carion2026sam3} that
was fine-tuned for this project with the text query ``landing pad'' and
then frozen (Appendix~\ref{app:provenance} describes what is known about
this fine-tuning). At inference it receives the same prompt and uses
confidence and mask thresholds of 0.5 at an internal resolution of 1008
pixels. DAM4SAM~\cite{videnovic2025dam4sam}, a tracker built on SAM~2.1
Hiera-Tiny~\cite{ravi2025sam2}, propagates the target when direct
detection fails, with its native object-presence processing enabled. A
CLIP ViT-B/32 image encoder~\cite{radford2021clip} supplies similarity
metadata.

A direct SAM~3 mask always takes precedence. Let $D_t$ be the direct mask
at perception step $t$ and $T_t$ the mask that the tracking session
propagates to the current frame. The mask passed to the actor is
\begin{equation}
    M_t =
    \begin{cases}
        D_t, & D_t \neq \emptyset, \\
        T_t, & D_t = \emptyset \text{ and } T_t \neq \emptyset, \\
        \mathbf{0}, & \text{otherwise}.
    \end{cases}
    \label{eq:mask_rule}
\end{equation}
In the base frontend, with which the actor was trained, a tracking
session is initialised on the first nonempty direct mask of an episode
and then advanced on every fresh frame. Later direct detections override
the output for their own frame but do not reinitialise the session, so
after a gap the tracker continues from its own accumulated memory. The
base frontend therefore already has temporal memory and is not a
detector-only baseline. Section~\ref{sec:temporal} introduces a frontend
that manages this memory differently.

On an NVIDIA L40S server GPU, one perception update took \SI{158.2}{ms} on
average with the base frontend and \SI{148.5}{ms} with Temporal Recovery
(Section~\ref{sec:temporal}), against the \SI{166.7}{ms} available at
\SI{6}{Hz}. Onboard latency was not measured.

\subsection{Observation, Action, and Policy}
\label{sec:policy}

\begin{figure}[!ht]
    \centering
    \includegraphics[width=\columnwidth]
    {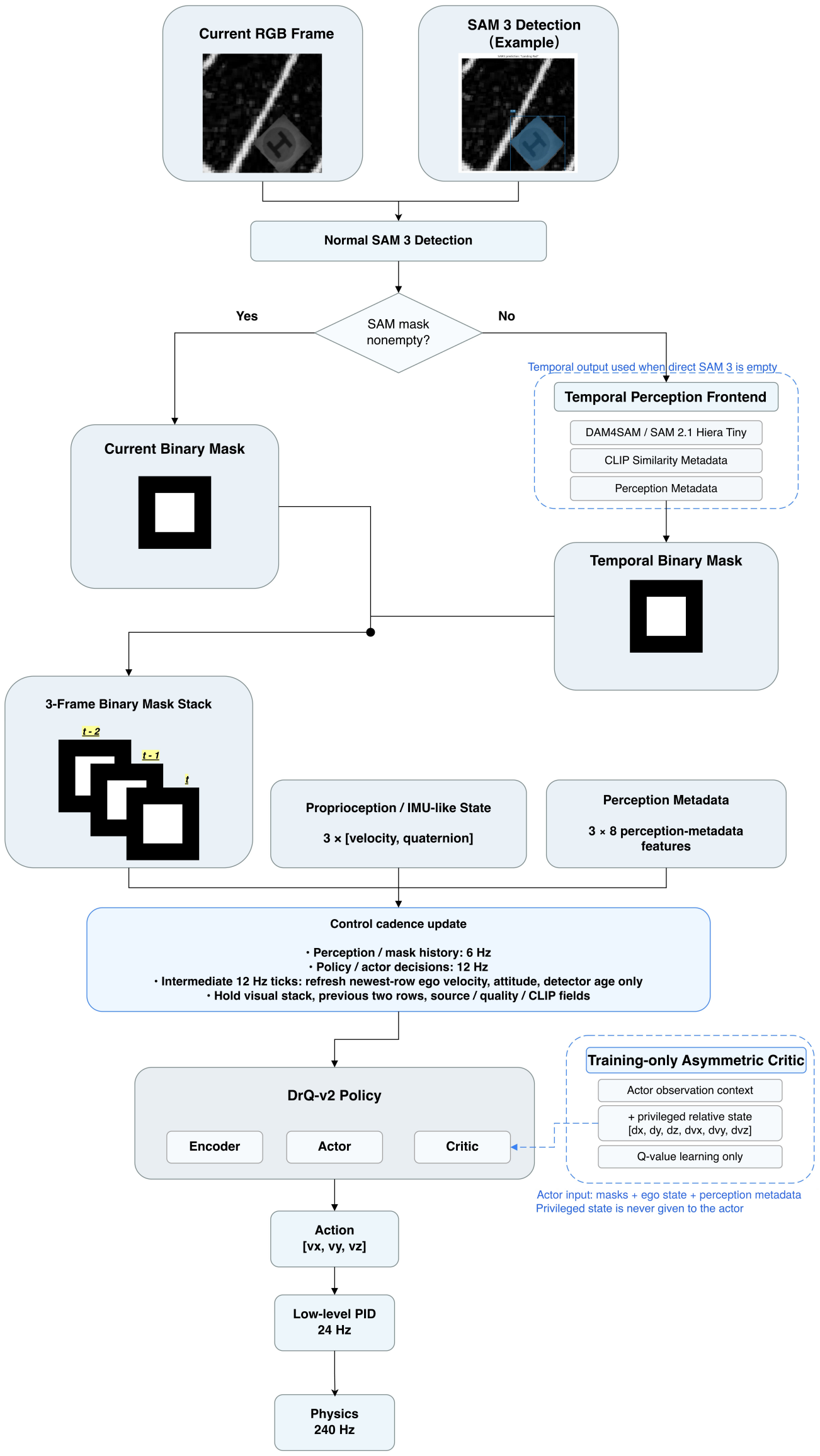}
    \caption{Overview of the shared perception, policy, and control
    pipeline. Direct SAM~3 masks take precedence and temporal predictions
    provide fallback observations. The evaluated frontends differ in their
    temporal anchoring (Section~\ref{sec:temporal}). CLIP similarity is
    part of the perception metadata and does not reject masks.}
    \label{fig:landing_pipeline}
\end{figure}

Figure~\ref{fig:landing_pipeline} summarises the pipeline. The actor
never receives the pose, velocity, or heading of the platform. Its
observation $o_t=(I_t,h_t)$ consists of the three most recent binary
masks,
\begin{equation}
    I_t = \left[ M_{t-2}, M_{t-1}, M_t \right]
    \in \{0,1\}^{3\times84\times84},
\end{equation}
and a low-dimensional history $h_t\in\mathbb{R}^{3\times15}$ with one row
per frame. Each row holds the world-frame velocity of the UAV (3 entries)
and its attitude quaternion (4), which stand in for onboard measurements.
It also holds eight perception metadata entries, computed only from the
images and the perception outputs. These are one-hot flags marking a direct, propagated, or
empty mask (3), the age of the last direct detection (1), the native
segmentation quality score and its availability (2), and a CLIP
similarity between the current mask crop and templates from earlier
direct detections, with its availability (2). The flags and the detection
age let the actor tell direct, propagated, and missing observations apart.
The actor outputs a world-frame velocity command
$a_t=[v^{\mathrm{cmd}}_{x,t},v^{\mathrm{cmd}}_{y,t},v^{\mathrm{cmd}}_{z,t}]^{\top}$,
limited to $\pm\SI{1.5}{m\,s^{-1}}$ horizontally and
$\pm\SI{0.5}{m\,s^{-1}}$ vertically.

The policy is trained with DrQ-v2~\cite{yarats2022drqv2}. The actor
encodes the mask stack with the DrQ-v2 convolutional encoder, concatenates
the flattened history, and maps the result through a 50-unit linear layer
with \texttt{BatchNorm1d} and tanh and two 512-unit hidden layers to a
bounded velocity command. Training uses an asymmetric
actor--critic~\cite{pinto2018}. Only the critic also receives the position
and velocity of the platform relative to the UAV. This privileged state
can ease value estimation, but it never enters the actor and is absent at
inference. The reward combines terminal rewards of $+140$ for
success, $-50$ for a crash, and $-45$ for a timeout with a small time cost
and potential-based shaping~\cite{ng1999} on the horizontal distance, the
horizontal relative speed, and the relative altitude, all computed from
simulator state. The actor was trained once, with one seed, for 1.25
million decisions, first under an easy-to-hard
curriculum~\cite{goldschmid2024,shin2026} and then under the nominal and a
speed-prioritised reset distribution (Appendix~\ref{app:provenance}).
Every comparison in Sections~\ref{sec:appearance} and~\ref{sec:temporal}
uses this frozen actor with exploration noise disabled.

\subsection{Evaluation}
\label{sec:protocol}

The closed-loop conditions are the nominal task distribution (P), an
exactly stationary platform (S0), and platform speeds restricted to
$[0,0.25)$, $[0.25,0.5)$, $[0.5,0.75)$, and $[0.75,1]$~\si{m\,s^{-1}}
(S1--S4). S0 is included because the nominal sampler produces an exactly
stationary platform with probability zero. In every comparison, the
evaluated configurations face the same episode seeds and reset states, so
outcomes are matched episode by episode. We report the success rate (SR)
together with crash and timeout counts, and compare matched success with
exact two-sided McNemar tests.

Two perception metrics describe what the actor receives. Terminal
usable-target coverage is the fraction of renderer-visible fresh camera
exposures in the final five simulated seconds of an episode for which the
returned mask is usable. These exposures are selected using their
recorded exposure-associated simulator step counters. A mask is usable if
it overlaps the rendered target by at least 20 pixels, with an
intersection over union (IoU) of at least 0.30 and a recall of at least
0.50, and an empty output on a visible target counts as unusable. The post-absence output rate is the fraction of fresh frames
after the last renderer-visible target frame of an episode on which the
frontend still returns a nonempty mask. Each configuration generates its
own trajectories, so both metrics are descriptive.

\section{Appearance Change and Prompt Re-Specification}
\label{sec:appearance}

\begin{table}[!t]
    \centering
    \small
    \caption{Appearance and prompt evaluation with the frozen actor and
    the Temporal Recovery frontend, 50 paired episodes per speed and
    condition. Entries are success/crash/timeout (success rate).
    A: original pad, original prompt; B: banana-themed pad, original
    prompt; C: banana-themed pad, re-specified prompt; D: sampled
    within-family greyscale pad texture, original prompt.}
    \label{tab:appearance}
    \footnotesize
    \setlength{\tabcolsep}{3pt}
    \begin{tabular}{ccccc}
        \toprule
        Speed & A & B & C & D \\
        (\si{m\,s^{-1}}) & & & & \\
        \midrule
        0.00 & 35/6/9 (70\%)  & 4/20/26 (8\%)  & 34/8/8 (68\%)  & 31/6/13 (62\%) \\
        0.25 & 38/2/10 (76\%) & 6/13/31 (12\%) & 34/8/8 (68\%)  & 42/3/5 (84\%) \\
        0.50 & 42/3/5 (84\%)  & 8/6/36 (16\%)  & 40/7/3 (80\%)  & 44/2/4 (88\%) \\
        0.75 & 42/1/7 (84\%)  & 6/4/40 (12\%)  & 37/6/7 (74\%)  & 41/3/6 (82\%) \\
        1.00 & 32/1/17 (64\%) & 6/2/42 (12\%)  & 26/4/20 (52\%) & 36/0/14 (72\%) \\
        \midrule
        All  & 189/13/48 & 30/45/175 & 171/33/46 & 194/14/42 \\
             & (75.6\%)  & (12.0\%)   & (68.4\%)  & (77.6\%) \\
        \bottomrule
    \end{tabular}
\end{table}

The appearance evaluation changes what the perception model has to
recognise while the actor, the perception networks, and the pad geometry
stay fixed. It uses the frozen actor with the Temporal Recovery frontend,
which reinitialises the tracker at each direct detection
(Section~\ref{sec:temporal}). Conditions A, B, and C share seeds and reset
states at each of five fixed platform speeds (0, 0.25, 0.5, 0.75, and
\SI{1}{m\,s^{-1}}), with 50 matched episodes per speed. Comparing A with B
isolates the appearance change under the original prompt ``landing pad'',
and comparing B with C isolates the prompt change on the same appearance.
The revised prompt, ``banana-printed landing pad'', is supplied only to
SAM~3; CLIP receives no text, although its image templates come from
the direct detections. The weights of the actor and of every perception
network are identical in all conditions. Condition D uses the seeds and
reset states of A but gives each of its 250 episodes a different pad
texture. The textures are sampled without replacement from a 424-image
greyscale library of variations within the landing-marker family of the
nominal pad.

The revised prompt came from exploratory testing. Several candidate
prompts were compared on fixed images of the banana-themed pad, without
running landing episodes and on samples separate from the evaluation
episodes. The evaluation episodes themselves had already been run once
with a memory-restricted variant of the frontend
(Appendix~\ref{app:memory}) before the evaluation reported here. The
result therefore shows prompt-mediated recovery in the tested setting, not
prospectively validated generalisation. Appendix~\ref{app:provenance}
gives the selection details.

Table~\ref{tab:appearance} reports the outcomes. Under the original
prompt, the banana-themed pad lowered success from 75.6\% to 12.0\%, and
most of the failures (175 of 220) were timeouts. Direct detections fell
from 63.3\% to 8.8\% of perception frames. The remaining frames shifted to
propagated masks, which rose from 21.3\% to 50.2\%, and to empty
observations, which rose from 15.4\% to 41.0\%. Whole-episode
usable-target coverage dropped from 92.8\% to 74.3\%. Changing only the text prompt brought direct detections back to
61.8\% of frames, coverage to 89.1\%, and success to 68.4\%. Both A and C
exceeded B at all five speeds. Pooled over the five speeds, 163 episodes
succeeded only in A and 4 only in B (exact two-sided McNemar test,
$p=3.4\times10^{-43}$), and 7 succeeded only in B and 148 only in C
($p=1.7\times10^{-35}$).

The recovery was incomplete. Success with the revised prompt stayed
below that on the original pad by 7.2 percentage points overall and by 2
to 12 points at each speed (57 A-only against 39 C-only successes,
$p=0.082$). Condition C also had 33 crashes against 13 in A, with a similar
number of timeouts (46 against 48). No degradation was detected with
within-family textures. Success was 77.6\% in D against 75.6\% in A (35
A-only against 40 D-only successes, $p=0.645$), per-speed differences
ranged from $-8$ to $+8$ points, and the perception statistics were
similar (direct 63.5\%, propagated 21.4\%, empty 15.0\%, coverage
93.1\%).

The binary mask keeps raw appearance out of the actor input, but the
perception model still has to detect the target. Landing success fell and
recovered together with the availability of direct detections, which
points to target-mask availability as an important limiting factor in
this setting. This is an association rather than an isolated effect. The
perception fractions come from different closed-loop trajectories, and
the experiment does not separate availability from other changes in the
observations, such as mask quality, metadata, temporal history, and the
resulting trajectories. The next section examines availability with a
controlled substitution.

\section{Target-Mask Availability Diagnosis}
\label{sec:diagnosis}

The appearance evaluation does not intervene on the masks themselves. To
identify which segmentation errors matter for a mask-conditioned
controller, we replace the observations of a frozen policy one error class
at a time. This requires a policy that can also act on renderer-derived
ground-truth masks. The diagnostic therefore uses a separate DrQ-v2 actor
trained on such masks under an earlier and simpler protocol, with shorter
episodes, lower starts, slower platforms, a smaller pad, and a different
reward and success criterion (Appendix~\ref{app:diagnostic}). The
raw-mask condition uses the same fine-tuned SAM~3 checkpoint and prompt as
the main study, without any temporal fallback. The diagnostic therefore
differs from the main study in the actor, the observation interface, and
the protocol. Its success rates cannot be compared with those of the main
actor, and its ground-truth condition is not an upper bound for that
actor. What it can show is which class of SAM error changes landing when
the controller is held fixed.

\begin{table}[t]
    \centering
    \caption{Frozen-policy substitution diagnostic with a separately
    trained ground-truth-mask actor, the same fine-tuned SAM~3 checkpoint
    as the main study, and the earlier protocol of
    Appendix~\ref{app:diagnostic}. Each repair modifies one class of SAM
    error; the actor is unchanged. These results are not comparable with
    those of the main actor.}
    \label{tab:frozen_diagnostic}
    \begin{tabular}{lcc}
        \toprule
        Observation condition & Screen & Confirmation \\
                              & (50 ep.) & (200 ep.) \\
        \midrule
        Ground-truth mask                     & 41/50 & 165 (82.5\%) \\
        Raw SAM-derived mask                  & 0/50  & 3 (1.5\%)  \\
        Centroid, area, or joint correction   & 0/50  & --     \\
        Visible-dropout oracle repair         & 42/50 & 174 (87.0\%) \\
        \bottomrule
    \end{tabular}
\end{table}

Table~\ref{tab:frozen_diagnostic} gives the results. In a 50-episode
screen, the ground-truth condition succeeded in 41 episodes and raw
SAM-derived masks in none. Correcting the centroid, the area, or both in
every nonempty SAM mask also gave 0 of 50. An oracle repair that replaces
an empty SAM output by the ground-truth mask only when the target is
visible in the rendered image, leaving every nonempty prediction
unchanged, gave 42 of 50. Only this repair was carried forward to a
200-episode confirmation. There, success was 82.5\% with ground-truth
masks, 1.5\% with raw SAM-derived masks, and 87.0\% with the repair.

The nonempty SAM masks were geometrically accurate. On the diagnostic
calibration set they had a mean IoU of 0.898, a mean centroid error of
0.70 pixels, and a mean area bias of $-7.9\%$. Their availability was the
weak point. Recall was 96.5\% when the target was fully visible, 46.7\%
when it was partially visible or at the image border, and 71.7\% over all
visible frames. Visible targets also tended to be missed in bursts, with a
median length of five perception frames. For the interventions tested,
correcting missing masks was much more consequential for landing than
correcting the tested geometric errors. The diagnostic covers only
centroid and area corrections and does not show that mask geometry is
irrelevant in general.

The main actor shows the same weak point in closed loop. It was trained on
SAM-derived masks, and with the base frontend its success is lowest on
stationary platforms (Section~\ref{sec:temporal}). We examined 27
terminal timeouts from a separate evaluation of this actor on slow or
stationary platforms, selected because the horizontal pad error had
fallen below \SI{1}{m}. In 25 of them, the target was still at least
partially visible at the first persistent loss of direct detection.
Prompting SAM~3 with an oracle target box, which is unavailable onboard,
recovered all 25 of these onset frames and 74 of 78 sampled frames in
which the target was visible but undetected. Detection thus failed while
the target could still be segmented. Later in these episodes,
renderer-confirmed target absence followed. In the final \SI{5}{s}, none
of the 805 perception calls without a direct detection contained
renderer-visible target pixels, and sustained renderer absence began a
median of \SI{2.08}{s} after the first persistent detection loss. This ordering is
observational and does not show that the detection loss caused the later
absence. It does show where target availability breaks down for the main
actor, and it motivates the question of Section~\ref{sec:temporal}:
whether the perception frontend can supply a usable target through these
gaps without retraining the actor.

\section{Temporal Memory Management}
\label{sec:temporal}

\subsection{Persistent Tracking and Temporal Recovery}
\label{sec:configs}

\begin{figure}[!ht]
    \centering
    \includegraphics[width=\columnwidth]
    {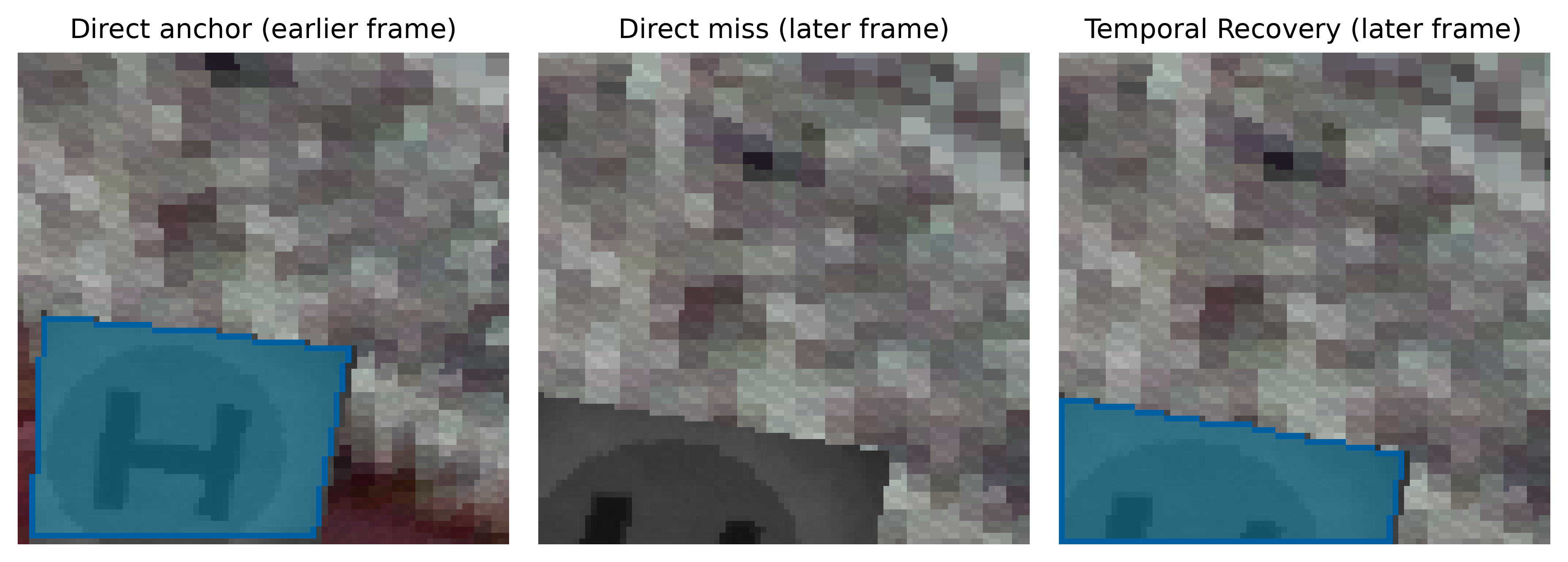}
    \caption{
    Example of temporal target recovery. A direct SAM~3 detection
    provides the earlier target observation (left). On a later frame
    without a direct detection (centre), the Temporal Recovery frontend
    propagates a target mask from the earlier visual evidence (right). The
    example is illustrative; closed-loop effects are reported in
    Section~\ref{sec:temporal_results}.
    }
    \label{fig:temporal_recovery_example}
\end{figure}

The base frontend bridges missed detections with one persistent tracking
session per episode (Section~\ref{sec:frontend}). After a long gap, that
session continues from memory accumulated over earlier propagated frames.
Temporal Recovery changes only how the session is anchored. With every
fresh nonempty direct SAM~3 mask, it reinitialises the tracking session on
the current frame, so that propagation always starts from the most recent
direct detection. The output rule of~\eqref{eq:mask_rule}, the observation
format, and every learned weight stay the same, and the actor receives no
tracker state, target position, or target velocity. The frontend is
causal, and an episode reset clears the session, the CLIP template bank,
and all counters. Figure~\ref{fig:temporal_recovery_example} shows an
example of a recovered target.

The Temporal Recovery wrapper also rejects a propagated mask whose native
object-presence logit is not finite and positive. For finite scores this
check repeats what DAM4SAM already does, because in the documented
implementation DAM4SAM empties the propagated mask whenever the score is
nonpositive. The effective difference from the base frontend is therefore
reinitialisation at direct detections. This conclusion rests on a review
of the source code (Appendix~\ref{app:provenance}); no closed-loop
ablation separated reinitialisation from the wrapper check. No predicted-IoU or CLIP threshold is applied to propagated
masks.

\subsection{Results}
\label{sec:temporal_results}

\begin{table}[!t]
    \centering
    \small
    \caption{Target continuity and post-absence output of the two
    frontends, pooled over the 300 episodes per frontend of the first
    evaluation (Table~\ref{tab:main_closed_loop}). Each frontend generates
    its own trajectories, so the values are descriptive.
    Terminal-coverage counts: 5{,}567/6{,}694 and 6{,}597/6{,}978
    renderer-visible fresh frames; post-absence output counts: 430/2401 and
    586/1945 fresh frames.}
    \label{tab:memory_tradeoff}
    \footnotesize
    \begin{tabular}{lcc}
        \toprule
        Metric & Base & Temporal Recovery \\
        \midrule
        Terminal usable-target coverage & 83.2\% & 94.5\% \\
        Post-absence output             & 17.9\% & 30.1\% \\
        \bottomrule
    \end{tabular}
\end{table}

Temporal Recovery made the target more continuously available. Its
terminal usable-target coverage exceeded that of the base frontend in all
six conditions of the first evaluation, by 5.0 to 16.6 percentage points,
and was 94.5\% against 83.2\% pooled over the evaluation
(Table~\ref{tab:memory_tradeoff}). The same reinitialised sessions also
kept returning masks more often after renderer-confirmed target absence.
After the last renderer-visible target frame of an episode, Temporal
Recovery returned a nonempty mask on 30.1\% of fresh frames, against
17.9\% for the base frontend. Renderer absence can arise from a lateral exit from the field of
view, clipping by the camera near plane, or other rendering effects, and
the recorded traces do not establish their relative contributions. These
rates are therefore the share of post-absence frames with a nonempty
output. They are not counts of false detections, and a nonempty output on
such a frame is not necessarily an error or a harmful control input.

\begin{table}[!t]
    \centering
    \small
    \caption{Landing outcomes of the frozen actor with the two frontends
    in the first evaluation, 50 matched episodes per condition. P and
    S1--S4 use separate seed blocks. Entries are success/crash/timeout
    (success rate). $p$: exact two-sided McNemar test, without
    multiplicity correction.}
    \label{tab:main_closed_loop}
    \footnotesize
    \setlength{\tabcolsep}{3pt}
    \begin{tabular}{lcccc}
        \toprule
        Condition & Base & Temporal Recovery & $\Delta$SR (pp) & $p$ \\
        \midrule
        P  & 44/1/5 (88\%)  & 48/1/1 (96\%) & $+8$  & 0.289 \\
        S0 & 30/9/11 (60\%) & 40/6/4 (80\%) & $+20$ & 0.076 \\
        S1 & 37/3/10 (74\%) & 36/5/9 (72\%) & $-2$  & 1.000 \\
        S2 & 42/4/4 (84\%)  & 45/1/4 (90\%) & $+6$  & 0.508 \\
        S3 & 47/1/2 (94\%)  & 44/2/4 (88\%) & $-6$  & 0.453 \\
        S4 & 37/3/10 (74\%) & 40/2/8 (80\%) & $+6$  & 0.549 \\
        \midrule
        Total & 237/21/42 (79.0\%) & 253/17/30 (84.3\%) & $+5.3$ & -- \\
        \bottomrule
    \end{tabular}
\end{table}

\begin{table}[!t]
    \centering
    \small
    \caption{Independent evaluation on 100 new matched episodes per
    condition. Entries are success/crash/timeout. The difference, its
    matched-pair 95\% confidence interval (Newcombe square-and-add method
    with Wilson limits, pointwise and approximate), and the exact two-sided McNemar $p$ come from
    a post hoc analysis that is conditional on the six unresolved pairs of
    Appendix~\ref{app:validity}; the original protocol withheld paired
    inference.}
    \label{tab:independent}
    \footnotesize
    \setlength{\tabcolsep}{2.5pt}
    \begin{tabular}{lccccc}
        \toprule
        Cond. & Base & Temporal Rec. & $\Delta$SR (pp) & 95\% CI (pp) & $p$ \\
        \midrule
        P  & 83/5/12 & 88/2/10  & $+5.0$ & $[-4.73,\,+14.74]$ & 0.405 \\
        S0 & 63/9/28 & 67/10/23 & $+4.0$ & $[-9.02,\,+16.83]$ & 0.652 \\
        \bottomrule
    \end{tabular}
\end{table}

Greater continuity did not translate into a reliable landing improvement.
In the first evaluation (Table~\ref{tab:main_closed_loop}), Temporal
Recovery landed in 253 of 300 episodes and the base frontend in 237.
Temporal Recovery was ahead in four of the six conditions and behind in S1
and S3. The largest difference was on stationary platforms (60\% against
80\%, $p=0.076$), and no per-condition difference reached the 0.05 level.
The independent evaluation (Table~\ref{tab:independent}) repeated the
comparison on new seeds. Temporal Recovery landed in 88 of 100 nominal
and 67 of 100 stationary episodes, against 83 and 63 for the base
frontend. These differences are smaller than in the first evaluation, so
the 48 of 50 nominal successes there are best read as a feature of that
sample.

To express the remaining uncertainty, we analysed all 200 matched pairs
of the independent evaluation post hoc. This analysis is conditional,
because six pairs show slight platform-position differences before
contact for reasons we could not identify (Appendix~\ref{app:validity}).
The differences of $+5.0$ and $+4.0$ percentage points have 95\%
confidence intervals of $[-4.73,+14.74]$ and $[-9.02,+16.83]$ points, so
the data remain compatible with both a benefit and a harm. Excluding the
six pairs leaves the direction unchanged, and the intervals still include
zero. Temporal Recovery thus landed numerically more often, but the
evaluations establish neither a reliable improvement nor equivalence or
the absence of a benefit.

\section{Discussion}
\label{sec:discussion}

The appearance evaluation shows that a frozen mask-conditioned landing
policy can recover most of its performance on a substantially different
pad when only the segmentation prompt is changed. This is adaptation
through the prompt, not invariance to appearance: under the original
prompt the banana-themed pad disrupted detection, and the policy landed
in 12.0\% of episodes. The mask interface moves the dependence on
appearance from the controller to the perception model, where a text
prompt can address it without new training data for the actor.

The results suggest that the availability of usable target masks was an
important limiting factor in the tested conditions. In the appearance
evaluation, landing success fell and recovered with the availability of
direct detections. The substitution diagnostic, which used a different
actor and protocol, points the same way. There, the nonempty masks were
already geometrically accurate, and restoring missing masks for visible
targets recovered landing, whereas the tested centroid and area
corrections did not. Frame-level segmentation accuracy is therefore a weak proxy for
the perception quality that this controller depends on, in line with
task-aware evaluation of perception in other domains~\cite{philion2020}.
Recall on visible targets, especially partially visible targets at the
image border, merits separate measurement when promptable segmentation is
used as a control interface.

More availability did not simply mean more landings. Temporal Recovery
raised terminal coverage, yet its effect on landing success remained
uncertain. Several explanations are possible, and our experiments cannot
distinguish between them. The base frontend already bridges gaps with a
persistent tracker, so the additional coverage may matter in few
episodes. Propagated masks carry their own source flag and may inform the
actor less than direct detections. The sessions that bridge detection gaps
also produce more nonempty output after renderer-confirmed target
absence, which may offset part of
the gain. Separating these effects would require configurations that were
not evaluated here, such as a frontend without temporal fallback, one that
holds the last direct mask, and a closed-loop contrast of
reinitialisation alone. Temporal memory management thus trades missing
observations against continued output after renderer-confirmed target
absence, which may carry outdated target information. The appropriate
degree of persistence depends on how the controller responds to each, which
segmentation accuracy alone cannot determine. A variant that restricts
propagation chains illustrates this trade-off (Appendix~\ref{app:memory}).

\subsection{Limitations}
\label{sec:limitations}

The appearance result concerns one substantially different pad, and the
revised prompt was chosen by the authors after exploratory screening on
development samples. Adaptation to arbitrary appearances and automatic
prompt selection were not tested. The within-family textures are not new
appearances, and the experiment does not separate the roles of colour,
contrast, and semantics. The detector is a SAM~3 model fine-tuned for this
project. Its original fine-tuning images were not available, so whether
banana or related artwork appeared in its fine-tuning data could not be
established (Appendix~\ref{app:provenance}), and its response to new
appearances and prompts reflects this model and not the public release.
Every frontend and appearance comparison uses one actor trained with one
seed. The substitution diagnostic uses a separate policy, protocol, and
observation interface, so its results do not transfer directly to the
main actor, and ground-truth masks were not supplied to the main actor.
The independent landing comparison rests on a post hoc analysis that is
conditional on six unresolved pairs. All results come from simulation,
with straight-line platform motion at constant speed and with ego velocity
and attitude provided by the simulator. The causes of renderer-confirmed
absence were not resolved, frontend latency was measured only on a server
GPU, and the pipeline has not been flown.

\section{Conclusion}
\label{sec:conclusion}

We studied prompt-conditioned segmentation as the observation interface
of a frozen visual RL policy for landing on a moving platform under a
change of pad appearance. Revising the segmentation prompt recovered
most of the landing performance lost under the tested appearance change
without updating any network weight. An important limiting factor was
whether the perception model delivered a target mask: landing followed the
availability of direct detections, and in a controlled diagnostic,
restoring missing masks mattered far more than correcting the tested
geometric errors. Temporal tracking can supply masks through detection
gaps, but it also increases nonempty output after renderer-confirmed
target absence, and a landing improvement from greater continuity was not
established. For
segmentation-conditioned RL landing, target-mask availability, and the
persistence with which a frontend supplies it, are the perception
properties to measure and design for, within the scope of these
simulation results.

\appendices

\section{Substitution Diagnostic Protocol}
\label{app:diagnostic}

The substitution diagnostic of Section~\ref{sec:diagnosis} used a DrQ-v2
actor trained on renderer-derived ground-truth masks and the earlier
protocol of Table~\ref{tab:diagnostic_protocol}. The confirmation used 200
paired episodes (seeds 2610000--2610199) after the 50-episode screen. The
saved checkpoint of the ground-truth-mask actor (Table~\ref{tab:provenance})
records a global step counter of 299{,}890, which counts outer
environment decisions and not physics steps, and a global episode counter
of 3{,}097. The intended total training budget, the training seed, and
the exact training-time source revision were not recovered. The raw-mask
condition used the fine-tuned SAM~3 checkpoint of the main study with the
prompt ``landing pad'', thresholds of 0.5, an internal resolution of 1008
pixels, and no temporal fallback.

\begin{table}[h]
    \centering
    \small
    \caption{Protocol of the substitution diagnostic, compared with the
    main protocol of Section~\ref{sec:system}.}
    \label{tab:diagnostic_protocol}
    \footnotesize
    \setlength{\tabcolsep}{3pt}
    \begin{tabular}{@{}lcc@{}}
        \toprule
        Setting & Diagnostic & Main \\
        \midrule
        Horizon / actor rate & \SI{24}{s} / \SI{6}{Hz} & \SI{30}{s} / \SI{12}{Hz} \\
        Initial altitude $h$ & 0.8--\SI{6}{m} & \SI{10}{m} \\
        Horizontal offset & radial, $\leq0.1\,h$ & $\mathcal{U}[-3,3]$~m/axis \\
        Platform heading & $\pm180^\circ$ & $\pm15^\circ$ \\
        Platform speed (\si{m\,s^{-1}}) & 0 (30\%), else $[0,0.5)$ & $\mathcal{U}[0,1]$ \\
        Pad size & \SI{0.675}{m} & \SI{1}{m} \\
        Horiz.\ velocity limit (\si{m\,s^{-1}}) & 1.42 & 1.5 \\
        Success criterion & altitude/contact & contact footprint \\
        Low-dim.\ values per frame & 7 & 15 \\
        SAM~3 checkpoint & \multicolumn{2}{c}{same fine-tuned checkpoint} \\
        Temporal fallback & none & frontend-specific \\
        \bottomrule
    \end{tabular}\\[1mm]
    \parbox{\columnwidth}{\footnotesize\raggedright
    The diagnostic also used an earlier reward. Its low-dimensional
    input held seven ego-state values per frame; the main actor receives
    seven ego-state values and eight perception-metadata values per row of
    $h_t$.}
\end{table}

\section{Independent Evaluation Statistics}
\label{app:validity}

\textit{Design.} The independent evaluation used seeds
22300000--22300099 for P and 22310000--22310099 for S0, each run once with
the base frontend and once with Temporal Recovery. All 400 episodes ran
exactly once, with no replacement, retry, or technical failure. The weights
of the frozen actor were verified to be identical in both arms.

\textit{Original pairing check and disposition.} Matched episodes had to
share their initial conditions, external platform motion commands, model
and configuration identities, and simulation timing, whereas UAV actions,
trajectories, outcomes, and durations may differ between frontends. The
check compared the world-frame centre-of-mass position of the pad link of
the platform vehicle between the two episodes. For each time-aligned
post-motion physics step $k$ in their common recorded prefix, the
difference is
\begin{equation}
    d_k = \max_{j\in\{x,y,z\}}
    \left|p_{\mathrm{Base},k,j}-p_{\mathrm{TR},k,j}\right|,
\end{equation}
the largest absolute coordinate difference, where TR denotes Temporal
Recovery. Before the first qualifying UAV--vehicle contact in either
episode, $d_k$ above \SI{1}{\micro m} required review; from that contact
onwards, differences were treated as possible physical responses. The
check classified 159 pairs as within the threshold throughout, 35 as
exceeding it only after contact, and 6 as exceeding it before contact (P:
78/17/5; S0: 81/18/1). Staying within the threshold does not mean exact
numerical equality. Table~\ref{tab:review_pairs} lists the six review
cases. In all six, the recorded exogenous conditions, including resets,
platform motion commands, model identities, and step and capture
schedules, matched exactly, and the recorded state does not reveal what
started the divergence. In five of them the threshold was first exceeded 2
to 15 physics steps before the first qualifying contact, so a near-contact
physical response is plausible but unproven. In the sixth, seed 22300048,
the first exceedance came about \SI{9}{s} before the recorded qualifying
contact, as isolated samples of at most \SI{1.43}{\micro m} rather than a
continuous excursion, with no established initiating cause. Because the original protocol required every
pair to pass the check, the independent evaluation was originally
reported with all 400 outcomes and without paired tests or confidence
intervals. These six cases remain unresolved and are not shown to be
harmless.

\begin{table}[h]
    \centering
    \small
    \caption{Independent-evaluation pairs with unresolved pre-contact
    divergence. One physics step is 1/240~s.}
    \label{tab:review_pairs}
    \footnotesize
    \setlength{\tabcolsep}{3pt}
    \begin{tabular}{lccc}
        \toprule
        Condition / seed & First step & First contact & Max.\ pre-contact \\
                         & $>$\SI{1}{\micro m} & step & difference (\si{\micro m}) \\
        \midrule
        P / 22300015  & 5000 & 5015 & 3.80 \\
        P / 22300036  & 5037 & 5039 & 1.49 \\
        P / 22300047  & 4915 & 4917 & 1.55 \\
        P / 22300048  & 4981 & 7196 & 1.43 \\
        P / 22300098  & 4968 & 4970 & 1.31 \\
        S0 / 22310096 & 4871 & 4874 & 1.35 \\
        \bottomrule
    \end{tabular}
\end{table}

\textit{Post hoc conditional reanalysis.} A post hoc analysis of the
existing records of all 200 matched pairs was later specified and frozen,
after the outcomes and the six-case review were known but before any test
or interval was computed. It is not a preregistered analysis and does not
retrospectively validate the original protocol. It rests on three
assumptions: the recorded matching adequately defines a common scenario
for each seed, unobserved state differences do not materially change the
scenario for one arm only, and the seeds act as independent scenario
replicates. Success differences were tested with exact two-sided McNemar
tests on the discordant pairs. Pointwise 95\% confidence intervals for the
matched-pair difference in success probability were computed with
Newcombe's square-and-add method for paired data~\cite{newcombe1998},
using Wilson intervals for the two marginal rates. These intervals are
approximate and are not inversions of the exact test.
Table~\ref{tab:posthoc} gives the results for all pairs and, as a
sensitivity analysis, with the six review pairs excluded. This exclusion
is itself post hoc and may introduce selection bias. With a Holm
adjustment over the two conditions, both all-pair $p$-values become
approximately 0.810. In every analysis the difference favours Temporal
Recovery, but every interval includes zero, so the data remain compatible
with both benefits and harms. The reanalysis therefore establishes neither an
improvement nor equivalence or the absence of a benefit.

\begin{table}[h]
    \centering
    \small
    \caption{Post hoc conditional reanalysis of the independent
    evaluation: Temporal Recovery minus base frontend. Disc.: discordant
    pairs, successes with the base frontend only / with Temporal Recovery
    only.}
    \label{tab:posthoc}
    \footnotesize
    \setlength{\tabcolsep}{2.5pt}
    \begin{tabular}{llccccc}
        \toprule
        Analysis & Cond. & Pairs & Disc. & $\Delta$SR (pp) & 95\% CI (pp) & $p$ \\
        \midrule
        All pairs      & P  & 100 & 9/14  & $+5.00$ & $[-4.73,\,+14.74]$ & 0.405 \\
                       & S0 & 100 & 20/24 & $+4.00$ & $[-9.02,\,+16.83]$ & 0.652 \\
        \midrule
        Six excluded   & P  & 95  & 9/12  & $+3.16$ & $[-6.70,\,+13.03]$ & 0.664 \\
                       & S0 & 99  & 19/24 & $+5.05$ & $[-7.97,\,+17.83]$ & 0.542 \\
        \bottomrule
    \end{tabular}
\end{table}

\textit{Other evaluations.} The first evaluation and the appearance
evaluation also use matched episodes, and their reset states were
verified, but the pre-contact platform-position check above was not
applied to them retrospectively. Two earlier runs of the base versus
Temporal Recovery comparison, a 40-episode pilot and a run stopped before
its planned sample was complete, are not analysed in this paper.

\section{Memory-Restricted Tracking}
\label{app:memory}

Memory-based video segmentation conditions later predictions on stored
temporal information, so an erroneous propagated mask can affect
subsequent frames~\cite{ding2025sam2long}. In Temporal Recovery, a session
reanchored at a direct detection keeps propagating through consecutive
misses, so its later predictions depend on earlier propagated masks. The
memory-restricted variant keeps as its anchor the RGB image and mask of
the most recent fresh direct detection, which only another direct
detection refreshes. On each fresh frame without a direct detection, a new
tracking session is initialised from this anchor and propagated once to
the current frame, so propagated predictions never become anchors for
later frames. The restriction is implemented by resetting sessions in the
wrapper; it does not reproduce the method of~\cite{ding2025sam2long}, no
model weight changes, and the observation format and cadence are those of
Temporal Recovery.

\textit{Screening.} Before any closed-loop evaluation, the variant was
screened against a previously set internal engineering criterion of at most six false
target reports on a fixed set of 54 identical frames, half of the 12
reports of Temporal Recovery on the same frames. The criterion was not a
statistical threshold or a physically derived limit. The variant produced
seven reports and did not meet it. We report its closed-loop results
below descriptively; the screening result alone does not determine their
interpretation.

\textit{Observations.} In the first evaluation
(Table~\ref{tab:memory_closed_loop}), the variant landed in 251 of 300
episodes, against 253 for Temporal Recovery and 237 for the base frontend
(Table~\ref{tab:main_closed_loop}), with a different profile across
speeds. Its terminal usable-target coverage was 90.8\% (6{,}321/6{,}958
renderer-visible fresh frames), between those of the base frontend
(83.2\%) and Temporal Recovery (94.5\%), and its post-absence output rate
was 12.8\% (238/1855 fresh frames), the lowest of the three. When
conditions A, B, and C of the appearance evaluation were run with this
variant on the same matched episodes, success was 78.0\%, 8.8\%, and
66.8\%, the same pattern as with Temporal Recovery; matched comparisons
with Temporal Recovery gave $p=0.561$, $0.280$, and $0.734$, which do not
show that the two frontends are equivalent. The variant illustrates the
persistence trade-off of Section~\ref{sec:discussion}: restricting
propagation lowered post-absence output at some cost in coverage. It is
not proposed as a method, and these results do not show a landing
improvement.

\begin{table}[h]
    \centering
    \small
    \caption{Landing outcomes of the memory-restricted variant in the
    first evaluation, 50 matched episodes per condition, matched with
    Table~\ref{tab:main_closed_loop}. Entries are success/crash/timeout
    (success rate).}
    \label{tab:memory_closed_loop}
    \footnotesize
    \begin{tabular}{lc}
        \toprule
        Condition & Memory-restricted \\
        \midrule
        P  & 41/2/7 (82\%) \\
        S0 & 44/1/5 (88\%) \\
        S1 & 43/2/5 (86\%) \\
        S2 & 38/5/7 (76\%) \\
        S3 & 44/1/5 (88\%) \\
        S4 & 41/2/7 (82\%) \\
        \midrule
        Total & 251/13/36 (83.7\%) \\
        \bottomrule
    \end{tabular}
\end{table}

\section{Implementation and Reproducibility}
\label{app:provenance}

\textit{Policy and training.} The DrQ-v2 encoder consists of four
unpadded $3\times3$ convolutions with 32 channels and strides of 2, 1, 1,
and 1, each followed by a ReLU, and yields 39{,}200 visual features, which
are concatenated with the 45 entries of the flattened history. Between
perception updates the mask history is held; on the intermediate actor
decision, the ego-state entries and the detection age of the newest row
are refreshed. The critic receives the platform position and velocity
minus those of the UAV in the world frame, normalised by $(3,3,10)$~m and
$(2.5,2.5,0.5)$~\si{m\,s^{-1}}. The encoder is updated only by critic
gradients, the actor receives its output detached, and the critic passes
the features and the six privileged scalars through its own trunk of the
same form and two Q heads with two 512-unit hidden layers each. The reward
is
\begin{equation}
    r_t = r_t^{\mathrm{term}}
        + \gamma\,\Phi(s_{t+1}) - \Phi(s_t)
        - 0.05\,\Delta t ,
\end{equation}
with the potential
\begin{equation}
\scalebox{0.9}{$
    \displaystyle \Phi(s) = -\Bigl[
        4\,\frac{e_{xy}}{e_{xy}+\SI{3}{m}}
      + 0.5\,\frac{v_{\mathrm{rel},xy}}{v_{\mathrm{rel},xy}+\SI{1}{m\,s^{-1}}}
      + \frac{z^{+}}{z^{+}+\SI{10}{m}}
    \Bigr]
$}
\end{equation}
of the horizontal distance $e_{xy}$, the horizontal relative speed
$v_{\mathrm{rel},xy}$, and the positive relative altitude $z^{+}$. For
terminal transitions $\Phi(s_{t+1})$ is set to zero. The decision interval
is normally $\Delta t=1/12$~\si{s}, but an early contact within the first
PID substep can give $1/24$~\si{s}. The discount
$\gamma\approx\sqrt{0.99}$ per actor transition keeps the physical-time
discount of an earlier \SI{6}{Hz} version of the actor, which used 0.99
per decision. Training ran for 500{,}000 curriculum decisions, starting
at \SI{2}{m} with offsets in $[-1,1]$~m and platform speeds in
$[0,1/3]$~\si{m\,s^{-1}} for the first 100{,}000 decisions, then at
\SI{5}{m} with $[-2,2]$~m and $[0,2/3]$~\si{m\,s^{-1}} until 250{,}000, and
then under the nominal distribution. A further 500{,}000 decisions used
the nominal distribution, and a final 250{,}000 sampled half of the
episodes from the nominal speed distribution and a quarter each from
$[0.5,0.75]$ and $[0.75,1]$~\si{m\,s^{-1}}. We did not establish what each
phase contributes. Table~\ref{tab:training_reward} lists the settings.

\begin{table}[h]
    \centering
    \small
    \caption{Training and reward settings of the main actor.}
    \label{tab:training_reward}
    \footnotesize
    \setlength{\tabcolsep}{4pt}
    \begin{tabular}{@{}ll@{}}
        \toprule
        Setting & Value \\
        \midrule
        \multicolumn{2}{@{}l}{Training (DrQ-v2)} \\
        Total actor decisions & $1.25\times10^{6}$ \\
        Mask stack & $3\times84\times84$ \\
        Low-dimensional history & $3\times15$ \\
        Optimiser / learning rate & Adam / $1\times10^{-4}$ \\
        Batch size / replay capacity & 256 / 160{,}000 \\
        Discount $\gamma^{\mathrm{e}}$ & 0.99498743710662 \\
        Target-update $\tau$ & 0.05 \\
        $n$-step return horizon / update interval & 6 / 2 decisions \\
        Random / update warm-up$^{\mathrm{a}}$ & 2{,}000 / 4{,}000 decisions \\
        Training seed & 1 \\
        Exploration $\sigma^{\mathrm{a}}$ & $1.0\rightarrow0.1$ over $5\times10^{5}$ \\
        Update noise clip & 0.3 \\
        Random-shift padding & 4 pixels \\
        Brightness multiplier$^{\mathrm{b}}$ & $\mathcal{U}(0.5,1.5)$ \\
        \midrule
        \multicolumn{2}{@{}l}{Reward} \\
        Success / crash / timeout$^{\mathrm{c}}$ & $+140$ / $-50$ / $-45$ \\
        Time cost$^{\mathrm{d}}$ & $-0.05\,\Delta t$ \\
        Horizontal-distance weight / scale & 4 / \SI{3}{m} \\
        Relative-horizontal-speed weight / scale & 0.5 / \SI{1}{m\,s^{-1}} \\
        Altitude weight / scale & 1 / \SI{10}{m} \\
        Terminal next-state potential & 0 \\
        \bottomrule
    \end{tabular}\\[1mm]
    \parbox{\columnwidth}{\footnotesize\raggedright
    $^{\mathrm{a}}$Counted in global actor decisions; not restarted during
    continuation.
    $^{\mathrm{b}}$Sampled per stack channel.
    $^{\mathrm{c}}$Terminal transitions also retain the time cost and the
    shaping term.
    $^{\mathrm{d}}$$\Delta t=1/12$~\si{s} normally; an early contact can
    yield $1/24$~\si{s}.
    $^{\mathrm{e}}$Per actor transition, $\approx\sqrt{0.99}$.}
\end{table}

\textit{SAM~3 fine-tuning.} The SAM~3 detector was fine-tuned for this
project in the run \texttt{landing\_pad\_generalized\_v2\_accum16}. The
archived fine-tuning pipeline specifies a single COCO category, ``landing
pad'', which is also the text query, and documents a randomised
landing-scene dataset with 400 training and 80 validation image entries,
of which 40 and 8 are intended as negative examples without a target.
These are documented dataset specifications, not verified counts of unique
images used in the final run: the original training images, masks, and
manifests were not available, and how the masks were produced could not
be established. The recorded configuration specifies 10 epochs, a model
input resolution of 1008 pixels, the AdamW optimiser, and frozen vision
and language backbones. Whether banana or related artwork appeared in the
fine-tuning data could not be established from the records. The first
author confirms that no banana-themed pad images or banana-related prompts
were deliberately included; this is an author statement, not a
verification against retained data. No statement is made about the
large-scale pretraining of the public SAM~3 model, whose exact upstream
release revision is not available. After fine-tuning, the checkpoint was
frozen during reinforcement learning and all evaluations.

\textit{Prompt selection.} Five candidate prompts were compared on the
same five fixed views of the banana-themed pad by scoring the resulting
SAM~3 masks against the rendered pad, without running any landing
episode: ``landing pad'', ``square landing pad'', and ``the full square
landing target'' returned no mask on any view, ``landing pad with banana
print'' gave two whole-pad masks, one partial mask, and two empty outputs,
and ``banana-printed landing pad'' gave four whole-pad masks and one empty
output. No prompt produced an acceptable mask on all five views, as the
screen required, and retests with the revised banana texture again gave
four acceptable views and one empty output. The prompt and texture were
then fixed, before any landing episode was run with them. With the prompt
fixed, the same conditions were first evaluated with the memory-restricted
frontend, on a block of 375 development episodes and then on a new set of
250 seeds. The Temporal Recovery evaluation of
Section~\ref{sec:appearance} came afterwards and reused this 250-seed set
to match the two frontends, so its episodes had already been evaluated
with the memory-restricted frontend. These seeds overlap neither the fixed
views of the prompt comparison nor the development episodes. The
available records do not indicate that the Temporal Recovery landing
outcomes were used to choose the prompt, but they do not exclude
undocumented prompt attempts.

\textit{Perception details.} The CLIP score is the mean cosine similarity
between the RGB bounding-box crop of the current output mask and the two
most similar templates in a causal bank of crops from direct detections,
or the similarity to the only template when the bank holds one; it uses
no text embedding. The rendered target mask used by the perception
metrics comes from the body identifier of the platform vehicle, including
its links. In the documented DAM4SAM implementation, a nonpositive
object-presence score replaces the predicted mask logits before
thresholding. In the independent evaluation, 2{,}881 perception frames
across 178 of the 200 Temporal Recovery episodes had a nonpositive
presence score, and every one produced an empty output. Latency was timed
synchronously on four recorded RGB sequences after 20 warm-up calls per
component; 2 of 585 base-frontend updates and 1 of 585 Temporal Recovery
updates exceeded the \SI{166.7}{ms} interval, and peak allocated GPU
memory was 4.60 and \SI{4.73}{GiB}.

\textit{Frozen components and runtime.} Table~\ref{tab:provenance} lists
the frozen components. The full SHA-256 digests of the SAM~3 and
ground-truth-mask actor checkpoints are
{\footnotesize\texttt{120dc5228a36cd1b1cb841a7607609c4}\allowbreak\texttt{b4f0cad30c1e3ad768cc8fd94cf30891}}
and
{\footnotesize\texttt{43bf505f11a5c96ced7b75fe384e6ec5}\allowbreak\texttt{e128185e9187ff65de2bede8d6e4ce70}}.
The recorded runtime environment used the optional compiled post-processing
fallback of SAM~2, and the independent evaluation logged a
texture-loading warning for both arms of seed 22310099 (S0).

\begin{table}[h]
    \centering
    \small
    \caption{Frozen components (first 12 hexadecimal characters of each
    SHA-256 digest).}
    \label{tab:provenance}
    \footnotesize
    \setlength{\tabcolsep}{3pt}
    \begin{tabular}{@{}lll@{}}
        \toprule
        Component & Configuration & SHA-256 \\
        \midrule
        Actor & 1.25\,M decisions, seed 1 & \texttt{926c61819980} \\
        SAM~3 & fine-tuned, 1008 px & \texttt{120dc5228a36} \\
        GT-mask actor & diagnostic only & \texttt{43bf505f11a5} \\
        DAM4SAM & \texttt{sam21pp\_hiera\_t.yaml} & \texttt{7402e0d864fa} \\
        CLIP & ViT-B/32 (\texttt{timm}, OpenAI) & \texttt{bd41409c7f2b} \\
        Temporal Recovery & runtime wrapper & \texttt{ccea0ddabc1c} \\
        \bottomrule
    \end{tabular}
\end{table}

\section*{Acknowledgment}

The first author used AI assistants during this work. ChatGPT and Codex
(OpenAI) mainly assisted with writing code, and Claude (Anthropic) helped polish its language and grammar. The
experiments, results, and scientific decisions are the
authors' own, and the authors checked all text, numbers, and references
and take full responsibility for the content of this paper.

\end{document}